\documentclass[conference]{IEEEtran}
\usepackage{graphicx,booktabs,amsmath,amssymb,xcolor,balance}
\usepackage{tikz}
\usepackage[hidelinks]{hyperref}
\usepackage{enumitem}

\newif\ifrevmarkup
\revmarkuptrue

\begin{document}

\title{Fleets Need a Context Plane: Rethinking Cooperative Perception for Autonomous Drones}

\author{
\IEEEauthorblockN{
Liangkai~Liu and
Xiaoxiao~Wu
}
\IEEEauthorblockA{
Department of Computer Science, Texas Tech University, Lubbock, USA
}
}

\maketitle
\thispagestyle{plain}\pagestyle{plain}

\begin{abstract}

Cooperative perception allows a drone fleet to combine observations from
multiple viewpoints. However, existing systems typically fix their
feature-sharing policies at design time or adapt to only one context
signal. This is a poor fit for aerial fleets, whose missions, bandwidth,
formation geometry, and scene coverage can change during flight. We
quantify the cost of context-blind sharing on UAV3D by controlling
feature exchange at evaluation time using a released DiscoNet
checkpoint, without retraining. Mission-aware sharing matches
full-sharing accuracy while using only 5--10\% of the bytes. The best
tested peer-selection policy changes with the byte budget, and choosing
the wrong policy loses up to 7.7~AP. Moreover, under a constrained
budget, two policies with the same full-scene accuracy differ by
5.9~AP within the mission region, showing that multiple context axes
must be considered jointly. We therefore propose the \emph{context
plane}, a bounded, structured interface for runtime context. Each drone
publishes a descriptor of at most 1\,KB at 10\,Hz, and lightweight,
replaceable policies use the fleet context to decide what each drone
computes, shares, and fuses. Existing sharing schemes become fixed
policies within this interface. In our ROS~2 prototype on a Jetson AGX
Orin, the context plane uses approximately 0.01\% of the data-plane
bandwidth, and each policy decision takes 0.10\,ms. These results show
that an explicit context interface can support low-overhead runtime
adaptation without modifying or retraining the perception model.

\end{abstract}

\section{Introduction}
\label{sec:intro}

Cooperative perception lets a drone fleet share sensor data so that
each drone sees around occlusions and from complementary
viewpoints~\cite{cooper, fcooper}. It is effective: on UAV3D, fusing
features across five drones doubles detection accuracy over a single
drone. Recent methods and benchmarks now bring
this capability from ground vehicles to aerial fleets~\cite{uav3d,
coped}, and they carry the sharing defaults of ground V2X with them.

SOTA systems decide \emph{what to share, with whom, and at what
fidelity and rate} at design time, and transmit to every peer, every
frame, at uniform fidelity~\cite{v2vnet, disconet, cobevt}. Some
methods adapt at runtime, but each responds to a single signal:
Where2comm sparsifies features by spatial
confidence~\cite{where2comm}, while HydraCollab adjusts fidelity to the
available bandwidth~\cite{hydracollab}. However, a drone fleet changes
along several axes at once within a single sortie: the mission switches
phase and each phase values different regions of the scene, the energy
budget is the flight time itself, the formation geometry changes in six
degrees of freedom, and the scene coverage of each drone overlaps and
separates. A policy that reads one of these signals, or none of them,
is left with the wrong sharing decision for most of the mission.

On UAV3D, using
the released DiscoNet~\cite{disconet} checkpoint with no retraining and
varying only the sharing schedule, we have three observations. First, full sharing is roughly an order of
magnitude too rich for the mission being flown: mission-relevant
accuracy matches it at 5--10\% of the bytes. Second, no fixed scheme
is safe, because the best sharing policy flips as the byte budget
moves, and holding the wrong one costs up to 7.7~AP within a single
sortie. Third, the context signals have to be taken together: at
starved budgets, the budget axis alone cannot choose between two
policies that mission context separates by about 6~AP. Existing
methods, including the most recent runtime adaptive one~\cite{hydracollab}, take at most one of these signals and embed the
response in model weights, where it cannot be reconfigured.

Considering these issues, we pose a critical question: \textit{Can
a drone fleet expose its operating context through a common interface,
so that sharing becomes a runtime decision rather than a
design-time static configuration?} We propose the \textbf{context plane}, a second
and inexpensive plane alongside the fleet's data plane of feature and
detection traffic. Each drone publishes its mission, scene, platform,
and geometry context in a single descriptor of at most 1\,KB at 10\,Hz
(Fig.~\ref{fig:arch}). Small and replaceable policies read the whole
fleet's context and schedule everything the data plane does: which
drone computes, which transmits, to which peers, and at what fidelity,
rate, and region. Policies never touch data-plane payloads, which keeps
every decision cheap to compute and easy to audit.

This work is timely for three reasons. First, aerial benchmarks
(UAV3D~\cite{uav3d}, CoPeD~\cite{coped}) have just appeared and port
the V2X protocol without modification. Second, Orin-class onboard compute makes
feature-level sharing feasible, so the bandwidth and energy
costs are immediate. Third, the transport already exists: DDS QoS and
ROS~2 topics can carry a context plane~\cite{ros2}, and only the
interface is missing. This paper makes four main contributions:
\begin{itemize}
\item \textbf{A context-plane abstraction for drone fleets.} The design defines a context descriptor, a plane contract, and a policy contract that enable context-aware adaptation of data sharing at runtime.

\item \textbf{An empirical study of context-blind sharing.} We quantify the benefits of using context to control data sharing. Adapting the sharing policy to context improves accuracy by up to 2.0 AP on held-out contexts compared with the best fixed policy. Jointly considering multiple context axes provides a further ${\sim}6$ AP improvement, without retraining the perception model.

\item \textbf{A ROS~2 prototype.} The context plane uses only ${\approx}0.01\%$ of the data-plane bandwidth it controls, and each policy evaluation takes 0.10,ms on a Jetson AGX Orin.

\item \textbf{A general evaluation methodology.} By adding evaluation-time context controls to released model checkpoints, our method can enable context-aware sharing policies for existing cooperative-perception models without retraining.
\end{itemize}

\section{Background and Motivation}
\label{sec:background}

\subsection{Cooperative Perception}

Cooperative-perception systems can share information at three
levels. Early fusion shares raw sensor data, preserving the most
information but requiring the most bandwidth~\cite{cooper}. Late fusion
shares only detection results, reducing bandwidth at the cost of
discarding useful information. Intermediate fusion shares latent
feature maps and balances these two extremes~\cite{fcooper}. Most
recent systems use intermediate fusion and focus on improving how
features are selected and combined. Examples include graph- and
attention-based fusion~\cite{v2vnet, disconet, v2xvit},
confidence-guided sparsification~\cite{where2comm}, and camera-only
collaboration that outperforms a single-agent LiDAR
baseline~\cite{coca3d}. Other work improves robustness to pose
errors~\cite{coalign} and heterogeneous fleets~\cite{heal}. However,
intermediate features remain large. On UAV3D, full sharing transmits
13.1,MB per fleet-frame (Section~\ref{sec:eval}), far exceeding the
capacity of a drone's wireless link. A practical system must therefore
decide \textit{what to share, with whom, at what fidelity, and at what
rate}.

\subsection{Motivation}

\noindent\textit{Sharing Decisions Are Fixed at Design Time.}
Most state-of-the-art systems set their sharing decisions before
deployment and keep them fixed throughout the mission. These choices
are largely inherited from ground-based V2X settings, where they are
often reasonable. Vehicles in common benchmarks~\cite{opv2v, v2xsim,
dairv2x} follow fixed routes, draw power from an engine or the grid,
maintain near-planar viewpoints, and rely on roadside infrastructure
and HD maps. A few systems adapt their sharing decisions at
runtime, but they typically respond to only one context
signal~\cite{where2comm, hydracollab}.

\vspace{0.5em}
\noindent\textit{Context Changes Rapidly in the Air.} These
assumptions do not hold for drone fleets. During a single flight, the
mission phase changes, batteries drain, formation geometry shifts, and
each drone observes different parts of the scene. Aerial deployments
may also lack supporting infrastructure. Therefore, a sharing policy
selected before takeoff is unlikely to remain effective throughout the
mission.

\section{System Design}
\label{sec:design}

\begin{figure}[t]
\centering
\begin{tikzpicture}[
  font=\scriptsize,
  box/.style={draw, rounded corners=1.5pt, align=center, inner sep=2.5pt},
  comp/.style={box, fill=black!6},
  drone/.style={},
]
\newcommand{\droneglyph}[3]{%
  \begin{scope}[shift={#1}, scale=#2]
    \draw[fill=black!12] (-0.16,-0.07) rectangle (0.16,0.07);
    \draw (-0.16,0.04) -- (-0.34,0.12); \draw (0.16,0.04) -- (0.34,0.12);
    \draw[fill=white] (-0.34,0.12) ellipse (0.11 and 0.028);
    \draw[fill=white] (0.34,0.12) ellipse (0.11 and 0.028);
    \node[below, yshift=-1pt] at (0,-0.07) {#3};
  \end{scope}}

\node[box, minimum width=3.15cm, minimum height=2.55cm, anchor=north west]
  (ego) at (-4.25, 1.62) {};
\droneglyph{(-3.55, 1.28)}{1.0}{}
\node[anchor=west, font=\scriptsize\bfseries] at (-3.1, 1.32) {ego drone $i$};
\node[comp, minimum width=1.25cm] (cam)  at (-3.55, 0.68) {cameras};
\node[comp, minimum width=1.25cm] (perc) at (-3.55, 0.08) {perception\\(BEV feat.)};
\node[comp, minimum width=1.25cm] (fuse) at (-3.55,-0.62) {fusion +\\detector};
\node[comp, minimum width=1.05cm] (pol)  at (-1.95, 0.38) {\textbf{policy} $\pi_i$};
\node[comp, minimum width=1.05cm] (gate) at (-1.95,-0.42) {gate};
\draw[->] (cam) -- (perc);
\draw[->] (perc) -- (fuse);
\draw[->] (pol) -- (gate) node[midway, right, xshift=1pt] {directives};
\draw[->] (perc.east) ++(0,-0.06) -| (gate.north west);

\droneglyph{(1.05, 1.30)}{1.0}{drone $j$}
\droneglyph{(2.75, 1.30)}{1.0}{drone $k$}

\draw[->, line width=2.2pt, black!35]
  (gate.east) -- (0.45, -0.42) -- (0.95, 0.95);
\draw[->, line width=2.2pt, black!35]
  (gate.east) -- (0.45, -0.42) -- (2.65, 0.95);
\draw[<-, line width=2.2pt, black!35]
  (fuse.south) ++(0.25,0) -- ++(0, -0.55) -- (3.3, -1.47) -- (3.3, 1.05);
\node[black!55, anchor=west] at (-2.45, -1.12)
  {\textbf{data plane}: features / detections (MB/s, gated)};

\draw[dashed, thick] (-4.1, 2.0) -- (3.4, 2.0);
\node[anchor=south west] at (-4.1, 2.02)
  {\textbf{context plane}: descriptors $\leq$1\,KB @ 10\,Hz};
\draw[dashed, ->] (-3.55, 1.62) -- (-3.55, 1.97);
\draw[dashed, ->] (1.05, 1.44) -- (1.05, 1.97);
\draw[dashed, ->] (2.75, 1.44) -- (2.75, 1.97);
\draw[dashed, ->] (-1.95, 1.97) -- (-1.95, 0.62)
  node[pos=0.12, right] {fleet ctx};

\node[box, minimum width=7.5cm, minimum height=0.34cm] at (-0.4, -1.72)
  {middleware: DDS / ROS~2 QoS};
\end{tikzpicture}
\caption{Context-plane design. Every drone publishes a $\leq$1\,KB
descriptor at 10\,Hz onto a fleet-wide context bus (dashed). Each
drone's policy reads \emph{fleet} context and emits directives; a gate
enforces them on the heavy data plane (gray pipes): which peers,
what representation, what ROI, what rate. Policies never touch
data-plane payloads.}
\label{fig:arch}
\end{figure}
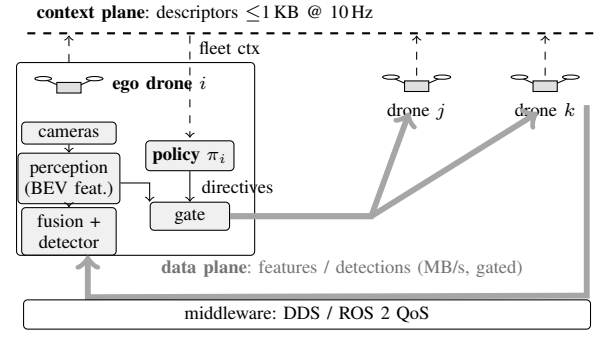

Fleets already have a \emph{data plane} that carries the heavy
perception payloads of cooperative perception, including feature maps,
detections, and tracks. However, they have no defined place for the
information that should govern this exchange. We propose an explicit
\textbf{context plane}, a low-rate and structured fleet-wide exchange of
operating context that small policies read to schedule the data plane.
Figure~\ref{fig:arch} shows the design. Every drone publishes a compact
context descriptor onto a fleet-wide bus. A policy on each drone reads
the context of the whole fleet and emits directives. A gate then
enforces those directives on the drone's data-plane traffic.

\subsection{Context descriptor}
Each agent publishes one compact descriptor at low rate ($\sim$10\,Hz,
$\leq$1\,KB), with four field groups; Table~\ref{tab:descriptor} gives
the concrete schema our prototype ships (180\,B serialized for a
four-peer fleet). Two design choices matter. First, the scene summary
is a \emph{coverage} statement, not content: a $16{\times}16$ BEV grid
at two bits per cell (seen well / seen poorly / unseen / novel) is
enough for peers to reason about which drone can contribute what
nobody else sees, without shipping features. Second, link quality is
reported \emph{per peer, as measured by the sender}, so every policy
sees the same (eventually consistent) link picture without a separate
monitoring service.

\begin{table}[t]
\centering
\caption{The context descriptor as implemented
(\texttt{ContextDescriptor.msg}); 180\,B serialized at $N{=}5$.}
\label{tab:descriptor}
\scriptsize
\setlength{\tabcolsep}{3.6pt}
\begin{tabular}{@{}llll@{}}
\toprule
\textbf{Group} & \textbf{Field} & \textbf{Type} & \textbf{Bytes} \\
\midrule
--- & stamp, agent\_id & Time, u8 & 9 \\
Geometry & position, orientation, velocity & f32[3,4,3] & 40 \\
 & footprint\_radius & f32 & 4 \\
Platform & battery, cpu, gpu headroom & f32$\times$3 & 12 \\
 & peer\_goodput[$N{-}1$] & f32[] & 16 \\
Scene & coverage\_grid ($16{\times}16\times$2\,b) & u8[64] & 64 \\
 & novelty\_flags & u8 & 1 \\
Mission & task\_phase & u8 & 1 \\
 & region\_of\_responsibility & f32[4] & 16 \\
 & priority\_target\_ids & u16[] & 2$k$ \\
\bottomrule
\end{tabular}
\end{table}

\subsection{The plane contract}

The plane contract follows six rules. First, each descriptor has a
bounded size and publication rate. Second, descriptors use best-effort
delivery and handle stale information conservatively. As a descriptor
ages, the policy may only reduce sharing fidelity or rate. In our
prototype, a descriptor older than three periods marks the link as
degraded, while one older than ten periods marks the peer as detached.
Third, the context plane carries only context descriptors, never
perception payloads. Fourth, policies read only these descriptors,
which keeps their decisions fast and auditable. Fifth, the system waits
before changing a policy, so small or temporary changes do not cause
rapid switching. Sending the same command again has no effect. This
prevents repeated commands which restarted the data timers and accidentally stop data transmission.
Finally, descriptors are authenticated
because they directly influence fleet-wide sharing decisions.

\subsection{The policy contract}
A policy is a small, swappable function
$\pi_i : \mathcal{C}^{N} \rightarrow \mathcal{D}^{N-1}$ from the
fleet's context (the $N$ latest descriptors, however stale) to one
directive per outgoing link. A directive includes six decisions:
(1)~whether to run expensive perception at all (duty cycle);
(2)~which peers to send to (topology); (3)~the representation on the
fidelity ladder, whose rungs span four orders of magnitude per link in UAV3D configuration: raw imagery 5.4\,MB/frame, full BEV features
655\,KB, budgeted features 164\,KB at $\rho{=}0.25$ and 13\,KB at
$\rho{=}0.02$, detections ${\sim}1$\,KB, tracks 0.24\,KB; (4)~the
spatial subset (ROI); (5)~rate and deadline; and (6)~receiver-side
fusion weights (trust, staleness). Policies may be engineered rules or learned
functions; the contract is agnostic. Because $\pi_i$ reads only local
copies of fleet context, no negotiation protocol is required. Policies
at different drones may transiently disagree, and the damping clause
plus age-monotonicity bound the cost of disagreement, which
disappears once the views converge.

\subsection{Unifying Existing Sharing Methods}

The context plane can represent the sharing mechanisms used by existing
cooperative-perception systems. Each method corresponds to a particular
policy in our design. Where2comm~\cite{where2comm} uses a scene
confidence map to select the ROI of shared features. When2com and
Who2com~\cite{when2com, who2com} use visual features as implicit scene
context to select communication peers. V2VNet~\cite{v2vnet} and
DiscoNet~\cite{disconet} use a fixed all-to-all topology and learn the
fusion weights. SyncNet~\cite{syncnet} uses network delay to compensate
for stale features at the receiver. Fixed-rate compression always uses
the same fidelity level. Concurrent work, HydraCollab
~\cite{hydracollab}, uses scene confidence to switch between
intermediate and late fusion.

Table~\ref{tab:subsumption} highlights two remaining gaps. First,
existing adaptive methods use at most one context axis and therefore
cannot jointly consider mission, scene, platform, and geometry.
Second, adaptation is often tied to a specific model architecture or
its weights, making it difficult to replace, audit, or verify. The
context plane does not replace these existing mechanisms. Instead, it
places them behind a common interface so that they can be selected and
combined according to runtime context.

\begin{table}[t]
\centering
\caption{Prior methods as fixed points in the context-plane policy
space. $\bullet$ = context read; decision columns list what the method
fixes at design time vs.\ leaves unaddressed.}
\label{tab:subsumption}
\scriptsize
\setlength{\tabcolsep}{2.6pt}
\begin{tabular}{@{}l cccc l l@{}}
\toprule
 & \multicolumn{4}{c}{\textbf{Context read}} & & \\
\cmidrule(lr){2-5}
\textbf{Method} & Mis. & Scene & Plat. & Geom. &
\textbf{Decides} & \textbf{Ignores} \\
\midrule
Where2comm~\cite{where2comm} & -- & $\bullet$ & -- & -- & ROI & topology, fidelity, rate \\
When2com~\cite{when2com} & -- & $\bullet$ & -- & -- & topology & ROI, fidelity, rate \\
Who2com~\cite{who2com} & -- & $\bullet$ & -- & -- & topology & ROI, fidelity, rate \\
V2VNet~\cite{v2vnet} & -- & -- & -- & -- & fusion wts & all scheduling \\
DiscoNet~\cite{disconet} & -- & -- & -- & -- & fusion wts & all scheduling \\
SyncNet~\cite{syncnet} & -- & -- & $\bullet$ & -- & fusion wts & topology, ROI, fidelity \\
HydraCollab~\cite{hydracollab} & -- & $\bullet$ & -- & -- & fidelity+ROI & topology, rate \\
Fixed compression & -- & -- & -- & -- & fidelity & everything else \\
\midrule
Context plane & $\bullet$ & $\bullet$ & $\bullet$ & $\bullet$ &
\multicolumn{2}{l}{all six, swappable at runtime} \\
\bottomrule
\end{tabular}
\end{table}

\subsection{A worked policy}
\label{sec:workedpolicy}

Our prototype implements the policy in Python. As summarized in Table~\ref{tab:policy}, the
battery level selects the initial representation and rate, while link
quality may reduce both. The mission phase determines the ROI, and
overlap in scene coverage selects the two most useful peers. We replayed our 120,s evaluation traces through this policy. As the
battery level decreases, the $1{\to}0$ link changes its representation
and rate at the 60\% and 30\% thresholds. When drone 2's link quality
drops at $t{=}60$,s, its outgoing links switch to detection-level
sharing within one descriptor period. None of these changes modifies
or retrains the perception model.

\begin{table}[t]
\centering
\caption{The prototype policy as a decision table (thresholds are
module constants; each rule reads only descriptor fields).}
\label{tab:policy}
\scriptsize
\setlength{\tabcolsep}{4pt}
\begin{tabular}{@{}lll@{}}
\toprule
\textbf{Context condition} & \textbf{Decision} & \textbf{Fields read} \\
\midrule
battery $\geq 60\%$ & feature @ 10\,Hz & platform \\
$30\% \leq$ battery $< 60\%$ & feature @ 5\,Hz & platform \\
battery $< 30\%$ & detections @ 2\,Hz & platform \\
peer goodput $< 5$\,Mbps & $\downarrow$ detections, rate $\leq 2$ & platform \\
phase $=$ track & ROI $=$ 32\,m disc & mission, geometry \\
always & top-2 peers by overlap & geometry \\
\bottomrule
\end{tabular}
\end{table}

\subsection{Middleware mapping and placement}
\label{sec:fixed}

The context plane sits between the perception stack and the
middleware. A descriptor publisher summarizes each drone's local
state, the policy node converts fleet context into directives, and the
gate applies those directives to data-plane publishers. These
components use standard DDS/ROS~2 QoS mechanisms~\cite{ros2} and
require no new transport.

Each drone publishes its descriptor on a best-effort, keep-last-1 topic
at 10,Hz. Message loss is handled by the staleness rules in the plane
contract. Directives use a reliable, keep-last-1 topic, and repeating
the same directive has no additional effect. Data-plane topics use QoS
profiles suited to their payloads: high-rate features use best-effort
delivery, while compact detections and tracks use reliable delivery.
The gate is the only component that reads both a directive and a
data-plane payload. Thus, the main contribution is not a new transport
protocol, but a standard interface for context-aware data sharing.

\subsection{Why not model-internal adaptation?}
\label{sec:whyplane}
One alternative is to place all adaptation inside a single fusion model
conditioned on battery level, mission phase, and formation geometry. We
instead place the adaptation logic behind a typed interface for four
reasons. First, drones from different vendors may use different
perception models, but they can still exchange the same context
descriptors and directives. Second, the logged context and policy
decisions clearly show why a drone changed or stopped sharing, while
model weights and attention scores are difficult to interpret. Third,
operators can change policies for different missions without retraining
the perception model. Fourth, a policy with bounded, typed inputs and
outputs is easier to check against mission-level safety requirements.
The interface does not exclude learned policies; a learned policy can
still implement the same contract.

\section{Implementation}
\label{sec:impl}

We implement and release two artifacts: a ROS~2 prototype of the
context plane and a measurement harness for evaluating context-aware
sharing on a public multi-UAV benchmark.

\subsection{Context-plane prototype}

The prototype consists of two ROS~2 Humble packages.
\texttt{context\_plane\_msgs} defines two message types.
\texttt{ContextDescriptor} contains the four context groups described
in Section~\ref{sec:design} and requires 180,B after serialization.
\texttt{DataPlaneDirective} specifies the source, destination,
representation, ROI, transmission rate, and deadline.

The \texttt{context\_plane} package provides four types of nodes. A
\emph{context publisher} runs on each drone at 10,Hz and publishes
either recorded or live platform state. A \emph{policy node} stores the
latest descriptor from every drone and maps the fleet context to
directives on each tick. Our battery-, link-, and phase-aware policy is
a 42-line Python function with no ROS dependencies, allowing it to be
unit-tested separately. A \emph{gate} on each drone applies the
directives by publishing synthetic, feature-sized payloads with the
specified representation, ROI fraction, and rate. Finally, a
\emph{benchmark node} counts the serialized bytes on every topic and
measures the CPU time used by the policy.

The policy reads only context descriptors. The gate is the only
component that reads both directives and data-plane payloads, preserving
the separation required by the contracts in
Section~\ref{sec:design}.

\subsection{Measurement harness}

The measurement harness evaluates cooperative-detection accuracy under
different context-driven sharing decisions. We use
UAV3D~\cite{uav3d}, a camera-only collaborative 3D-detection benchmark
with five UAVs, and the released DiscoNet~\cite{disconet} checkpoint
without retraining.

All sharing decisions are applied at the feature-exchange point during
evaluation. Each peer produces an 80-channel FP16 BEV feature map. A
budget $\rho$ limits the total number of cells shared with each ego
drone to a fraction $\rho$ of full sharing. A peer-selection policy
chooses which drones contribute features, and an allocation rule divides
the budget among them. The allocation can be uniform or proportional
to footprint overlap computed from the drone poses. We calculate the
transmitted bytes exactly as
\begin{align}
\text{retained cells} \times 80\text{ channels} \times 2\text{ B}.
\end{align}

The harness records per-frame predictions, ground truth, and
per-agent communication costs. A separate evaluator computes rotated
BEV IoU and AP at an IoU threshold of 0.5. All accuracy results in this
paper use this evaluator and the same evaluation protocol, ensuring
consistent comparisons. The third-party DiscoNet model is used only
for inference; the sharing policies, byte accounting, and evaluation
are implemented and tested in our code.

Applying the context controls at evaluation time also provides a
general measurement method. A released cooperative-perception
checkpoint can serve as a probe of the policy space without retraining,
while the harness records the exact communication cost of each policy.

\begin{figure*}[t]
\centering
\includegraphics[width=\textwidth]{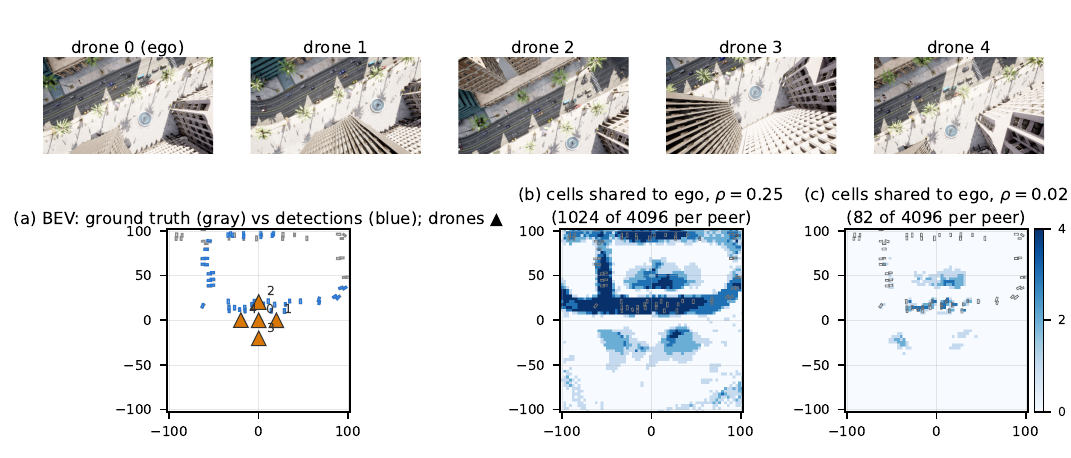}
\caption{One UAV3D mini-validation frame processed by the measurement
harness. Top: the five drones' nadir camera views.
(a)~BEV ground truth (gray) and DiscoNet detections (blue);
$\blacktriangle$ marks the cross-shaped drone formation.
(b,c)~The feature cells shared by the four peers with the ego drone at
$\rho{=}0.25$ and $\rho{=}0.02$. Shading indicates how many peers share
each cell, and gray boxes show the ground truth. At the lower budget,
the selected cells concentrate around the strongest object evidence.}
\label{fig:data}
\end{figure*}

Figure~\ref{fig:data} shows one mini-validation frame passing through
the harness. It includes the five camera views, the fused BEV
detections and ground truth, and the exact feature cells transmitted by
each peer under two budgets. At $\rho{=}0.25$, the retained cells follow
the object structure of the scene. At the more limited
$\rho{=}0.02$ budget, sharing concentrates on the strongest object
evidence.

\section{Evaluation}
\label{sec:eval}

Our evaluation covers two aspects. First, how much accuracy is lost
when a fixed sharing policy ignores runtime context? Second, the computation and communication costs of the context plane.

We evaluate context-blindness on the 80-frame UAV3D mini-validation
split, which contains one frame from each scene. All policies are
compared under the same byte budgets and evaluation protocol. Full
sharing transmits 13.1\,MB per fleet-frame. Because we use the
mini-validation split, the absolute AP values should not be compared
with results on the full test split. We measure system overhead on a
Jetson AGX Orin.

Unless stated otherwise, brackets report 95\% paired-bootstrap
confidence intervals over the 80 frames. Because each frame comes from
a different scene, we treat the frames as independent samples.
``n.s.'' indicates that the confidence interval contains zero.

Our evaluation uses one benchmark, UAV3D, and one released model,
DiscoNet. These experiments establish that context-blind sharing has a
measurable cost and that using context can recover accuracy. We do not
claim that the exact values will remain unchanged across other
datasets, fusion models, or wireless networks. Our broader claim is
that the best sharing policy depends on runtime context and that no
single fixed policy performs best across the full operating range.
Evaluation across additional models, datasets, and real wireless
conditions remains future work.

\subsection{Cooperative gain and the byte-accuracy frontier}
\label{sec:eval-coop}

Without fusion, a single drone achieves 0.185 AP@0.5. Full cooperative
sharing increases accuracy to 0.371, a gain of 18.5~AP
$[+16.9,+20.3]$. Cooperation is therefore clearly beneficial. However, the byte--accuracy curve in Fig.~\ref{fig:tax}a quickly
saturates. Reducing the shared bytes by $10\times$, to
$\rho{=}0.1$, costs only 2.0~AP $[+0.8,+3.4]$ relative to full sharing.
At $\rho{=}0.05$, the system retains 58\% of the cooperative gain while
using only 5\% of the bytes. In contrast, reducing the budget to
$\rho{=}0.01$ lowers accuracy by 24~AP. Thus, the last order of
magnitude in bandwidth provides only about two additional points of AP.


\subsection{Mission context}
We next evaluate accuracy only within a 32\,m mission region,
representing a track-a-target task. Mission-specific AP reaches
0.773--0.784 for $\rho\in[0.05,0.1]$, which is statistically
indistinguishable from the 0.763 AP achieved with full sharing. The
largest difference is ${+}2.2$~AP $[-0.5,+5.1]$ (n.s.;
Fig.~\ref{fig:tax}a).

Therefore, a mission-aware policy can reduce communication by
$10\times$ without a measurable loss in task-relevant accuracy.
Context-blind full sharing instead spends bandwidth and energy on
regions that the current mission does not prioritize.

\subsection{Peer selection}

We compare peer-selection policies at a fixed budget of
$\rho{=}0.25$. Every configuration transmits the same total number of
bytes. Among the two-peer policies, selecting the two most-overlapping
peers (\emph{best-2}), the two least-overlapping peers
(\emph{worst-2}), or two random peers produces no significant
difference:
[
\text{worst-2} - \text{best-2}
= {+}0.8~\text{AP};[-0.8,+2.3],
]
and
[
\text{worst-2} - \text{random-2}
= {+}0.1~\text{AP};[-1.1,+1.3].
]
Neither difference is significant. The important factor at this budget is spreading the bytes across more
peers. Sharing through all four peers outperforms every two-peer policy
by ${+}4.8$~AP $[+3.8,+5.8]$. In this dense aerial formation,
pairwise overlap does not reliably predict peer value. Effective peer
selection therefore requires the coverage context carried by the context
descriptor, rather than a fixed geometric
rule.

\subsection{Budget sensitivity}
\label{sec:eval-flip}
\begin{figure}[t]
\centering
\includegraphics[width=\linewidth]{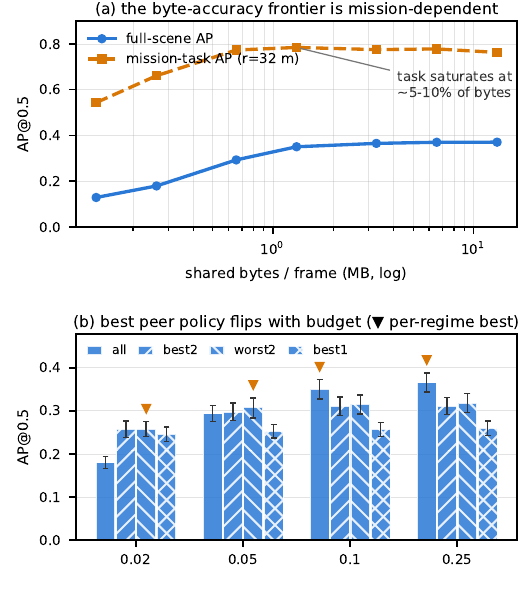}
\caption{The context-blindness tax on UAV3D (DiscoNet, released
checkpoint, eval-time knobs, matched bytes within every comparison).
(a)~Full-scene AP saturates while mission-task AP peaks at 5--10\% of
full-sharing bytes. (b)~At matched bytes, the best peer-selection
policy flips between budget regimes ($\blacktriangledown$ marks each
regime's best; whiskers are 95\% bootstrap CIs); a context-switched
oracle beats the best fixed policy by 2.1 mean-AP (2.0 held-out), and
the wrong fixed choice pays up to 7.7~AP in-regime.}
\label{fig:tax}
\end{figure}

We repeat the matched-byte comparison across four budgets
(Fig.~\ref{fig:tax}b). At the smallest budget, $\rho{=}0.02$,
concentrating the available bytes on two peers outperforms spreading
them across all four peers by ${+}7.7$~AP $[+6.6,+8.9]$.
Concentration still leads by ${+}1.4$~AP $[+0.2,+2.7]$ at
$\rho{=}0.05$.

The result reverses as the budget increases. At $\rho{=}0.1$, spreading
across all four peers outperforms the best two-peer policy by
${+}4.0$~AP $[+2.8,+5.3]$. At $\rho{=}0.25$, the advantage increases
to ${+}4.8$~AP $[+3.8,+5.8]$. The crossover occurs between 5\% and
10\% of full-sharing bandwidth, a range that a real wireless link can
easily cross during operation.

No fixed policy performs well across the entire range. The policy that
wins with a large budget loses 7.7~AP when the link is starved, while
the policy that wins with a small budget loses 4.8~AP when bandwidth is
plentiful.

To measure the value of context-dependent selection, we construct an
oracle that selects the best evaluated fixed policy for each budget.
Across the four budgets, the oracle achieves 0.320 mean AP, compared
with 0.299 for the best single fixed policy, an improvement of
${+}2.1$~AP. We also select the winning policy using half of the frames
and evaluate it on the remaining half. The held-out improvement remains
${+}2.0$~AP, showing that the gain is not caused by selection bias.

This oracle is a measurement tool, not an online policy. It only
switches among the fixed policies already evaluated and does not
explore the larger directive space supported by the context plane.
Designing practical online policies over that space, using either rules
or learning, remains future work.

In the evaluated cross formation, drones are separated by approximately
20\,m and have 42\,m footprints. Because their overlaps are similar,
overlap-proportional allocation performs about the same as uniform
allocation. This result does not imply that allocation is unimportant;
its effect should become stronger as the formation spreads and overlap
becomes less uniform.

\subsection{Composition of context axes}
\label{sec:eval-compose}

The context plane is designed to combine multiple context axes, and the
results show why this is necessary. At $\rho{=}0.02$, the two
concentration policies, best-2 and worst-2, both achieve 0.257 AP on
the full scene. The budget alone therefore cannot determine which two
peers should be selected. Adding mission context breaks this tie. For the track-a-target task,
best-2 achieves 0.787 AP, compared with 0.728 AP for worst-2, a
difference of ${+}5.9$~AP $[+2.5,+9.4]$. The mission and budget axes
also interact at $\rho{=}0.05$. A concentration policy performs best
on the full scene, while spreading performs better within the mission
region by ${+}3.1$~AP $[-0.6,+7.0]$. This second result is
directionally consistent but not statistically significant at the
current sample size.

Overall, a policy that reads only the budget leaves approximately
6~AP unclaimed compared with one that jointly reads budget and mission
context. Both signals fit within the same 1\,KB descriptor.

\subsection{Overhead of the context plane}
\label{sec:eval-overhead}
Four simulated agents replay 120\,s context traces containing a mission
phase change, battery decay, and a link-quality drop. We run the
prototype natively on a Jetson AGX Orin with 64\,GB of memory, a
companion-computer platform suitable for capable drones. Four 180\,B descriptors published at 10\,Hz, together with 80
directives per second, use 10.1\,KB/s. The data plane controlled by
these messages transmits 105\,MB/s. The context plane therefore adds
approximately 0.01\% of the data-plane bandwidth.

Policy evaluation takes 0.10\,ms at the median and 0.13\,ms at the
99th percentile. This is approximately three orders of magnitude
below the 100\,ms deadline imposed by the 10\,Hz update rate. The full
ROS~2 prototype uses 9\% of one CPU core. An x86 container produces
similar results, with a 62\,MB/s data-plane rate and a median policy
latency of 0.03\,ms. The replayed traces exercise the complete fidelity ladder. At
$t{=}45$\,s, a mission-phase change reduces feature payloads by
$10\times$ through an ROI directive. Battery decay moves agent~1 down
the rate ladder, and a link drop at $t{=}60$\,s switches agent~2 to
detection-level sharing. Each transition is triggered by context, and
none modifies the perception model.

\section{Discussion}
\label{sec:discussion}


The reorganization makes five problems addressable.
\textit{(1)~Policy synthesis and verification.} Policies are
bounded functions of bounded typed inputs, so mission-level safety
properties (``never drop detection sharing while a human is in
the region of responsibility'') may be checkable mechanically.
\textit{(2)~Context-plane security.} A drone that lies about
its coverage can stop its peers from looking at a region, so trust
calibration for small and cheaply cross-validated descriptors deserves
study before the interface is standardized.
\textit{(3)~Active cooperation.} We scoped this paper to
passive sharing, but drones also choose their viewpoints. Mission and
coverage context together form a natural basis for deciding which drone
repositions to see what, which is the extension with the largest payoff
and the largest safety obligations.
\textit{(4)~Context-conditioned benchmarks.} Aerial
datasets~\cite{uav3d, coped} ship frames but no battery, link, or
mission traces, so policies cannot be evaluated where they matter.
We release our traces and harness as a start.
\textit{(5)~Minimal sufficient context.} The smallest
descriptor that preserves policy quality is a rate-distortion
question, and its answer sets how cheap the plane can ultimately
be.

\section{Related Work}
\label{sec:related}

\noindent\textbf{Collaborative perception.}
Section~\ref{sec:fixed} shows how existing collaborative-perception
methods~\cite{cooper, fcooper, where2comm, when2com, who2com, v2vnet,
disconet, syncnet, hydracollab} fit within our policy space. The closest
recent method is HydraCollab~\cite{hydracollab}, which switches between
fidelity levels at runtime. However, it responds only to scene
confidence and does not combine multiple context axes. 

Other work improves the fusion model itself, including transformer
fusion~\cite{v2xvit, cobevt} and camera-only
collaboration~\cite{coca3d}. These methods determine how received
features are combined, while the context plane determines which
features should be sent, by which drones, and at what fidelity and
rate. The two directions are therefore complementary. The same
distinction applies to methods that handle pose errors~\cite{coalign}
or heterogeneous agents~\cite{heal}: they improve the fusion of
received data, whereas the context plane controls the data that
arrive. Finally, aerial and multi-robot datasets~\cite{uav3d, coped}
provide perception data but generally lack operational context such as
battery, link-quality, and mission traces. We release our traces to
help fill this gap.

\vspace{2mm}
\noindent\textbf{Context and control planes in other systems.}
Our system follows established ideas rather than claiming the
general concept of a separate plane. Examples include Clark et al.'s
\emph{knowledge plane}~\cite{knowledgeplane}, the separation of control
and data in software-defined networking, context-aware
computing~\cite{dey2001context}, and the fidelity ladder used in
adaptive bitrate streaming. Task-oriented
communication~\cite{taskoriented} is especially relevant because it
argues that a link should carry information needed by the current task.
Our mission-context results apply this principle to cooperative BEV
perception.

To our knowledge, prior work does not provide a fleet-wide bus to which
every drone publishes context and from which policies derive
perception-specific directives, such as peer selection, representation,
ROI, rate, and fusion trust. This interface must also account for
flight-time energy constraints and support auditable decisions.


\section{Conclusion}
\label{sec:conclusion}


Fixed sharing is a poor mechanism for aerial cooperative perception
because mission goals, bandwidth, battery level, formation geometry,
and scene coverage all change during flight. We introduce the \emph{context plane}, which separates lightweight operating context from large perception payloads and allows policies to adapt what each drone computes, shares, and fuses. Our UAV3D evaluation shows that mission-aware sharing preserves task-relevant accuracy with only 5--10\% of full-sharing bytes, while the best sharing policy changes with the available bandwidth. Pairwise geometry alone cannot reliably identify the most useful peers, and combining multiple context axes provides approximately 6~AP over using a single axis. Our ROS~2 prototype adds only 0.01\% communication overhead and takes 0.10\,ms per policy decision on a Jetson AGX Orin. Together, these results show that the context plane can adapt sharing decisions to runtime conditions with minimal overhead.

\balance
\bibliographystyle{IEEEtran}
\bibliography{refs}

\end{document}